\documentclass[10pt,twocolumn,letterpaper]{article}

\usepackage[accsupp]{axessibility}  %

\usepackage{cvpr}              %

\usepackage{graphicx}
\usepackage{amsmath}
\usepackage{amssymb}
\usepackage{booktabs}
\usepackage{multibib}
\usepackage{caption}
\usepackage{subcaption}
\usepackage{multirow}
\usepackage{url}            %
\usepackage[table]{xcolor}
\usepackage[british, english, american]{babel}
\usepackage[pagebackref=false, breaklinks=true, letterpaper=true, colorlinks, urlcolor=gray,
            citecolor=citecolor, linkcolor=linkcolor, bookmarks=false]{hyperref}
\definecolor{citecolor}{HTML}{0071BC}
\definecolor{linkcolor}{HTML}{ED1C24}

\newcites{app}{Appendix References}

\newlength\savewidth\newcommand\shline{\noalign{\global\savewidth\arrayrulewidth
  \global\arrayrulewidth 1pt}\hline\noalign{\global\arrayrulewidth\savewidth}}
\newcommand{\tablestyle}[2]{\setlength{\tabcolsep}{#1}\renewcommand{\arraystretch}{#2}\centering\footnotesize}

\makeatletter\renewcommand\paragraph{\@startsection{paragraph}{4}{\z@}
  {.5em \@plus1ex \@minus.2ex}{-.5em}{\normalfont\normalsize\bfseries}}\makeatother

\newcolumntype{x}[1]{>{\centering\arraybackslash}p{#1pt}}
\newcolumntype{y}[1]{>{\raggedright\arraybackslash}p{#1pt}}
\newcolumntype{z}[1]{>{\raggedleft\arraybackslash}p{#1pt}}

\newcommand{\app}{\raise.17ex\hbox{$\scriptstyle\sim$}}
\newcommand{\x}{{\times}}
\definecolor{baselinecolor}{gray}{.9}

\newcolumntype{*}{>{\global\let\currentrowstyle\relax}}
\newcolumntype{^}{>{\currentrowstyle}}
\newcommand{\rowstyle}[1]{\gdef\currentrowstyle{#1}#1\ignorespaces}

\definecolor{dt}{gray}{0.7}  %

\usepackage[capitalize]{cleveref}
\crefname{section}{Sec.}{Secs.}
\Crefname{section}{Section}{Sections}
\Crefname{table}{Table}{Tables}
\crefname{table}{Tab.}{Tabs.}

\def\cvprPaperID{1236} %
\def\confName{CVPR}
\def\confYear{2023}

\begin{document}

\title{Scaling Language-Image Pre-training via Masking}

\author{
Yanghao Li$^*$ \quad Haoqi Fan$^*$ \quad Ronghang Hu$^*$ \quad Christoph Feichtenhofer$^{\dagger}$ \quad Kaiming He$^{\dagger}$ \\[2mm]
 \small $^*$equal technical contribution, $^{\dagger}$equal advising \\[2mm]
 Meta AI, FAIR
  \vspace{5pt}\\
  \hypersetup{urlcolor=magenta}\url{https://github.com/facebookresearch/flip}
\vspace{-1em}
}
\maketitle

\begin{abstract}
	\vspace{-5pt}
We present Fast Language-Image Pre-training (FLIP), a simple and more efficient method for training
CLIP \cite{Radford2021}. Our method randomly masks out and removes a large portion of image patches during training. Masking allows us to learn from more image-text pairs given the same wall-clock time and contrast more samples per iteration with similar memory footprint. It leads to a favorable trade-off between accuracy and training time. In our experiments on 400 million image-text pairs, FLIP improves both accuracy and speed over the \mbox{no-masking} baseline. On a large diversity of downstream tasks, FLIP dominantly outperforms the CLIP counterparts trained on the same data. 
Facilitated by the speedup, we explore the scaling behavior of increasing the model size, data size, or training length, and report encouraging results and comparisons. 
We hope that our work will foster future research on scaling vision-language learning.    
\vspace{-1em}
\end{abstract}

\vspace{-10pt}
\section{Introduction}
\label{sec:intro}

\begin{figure}[t]\centering
\includegraphics[width=0.95\linewidth]{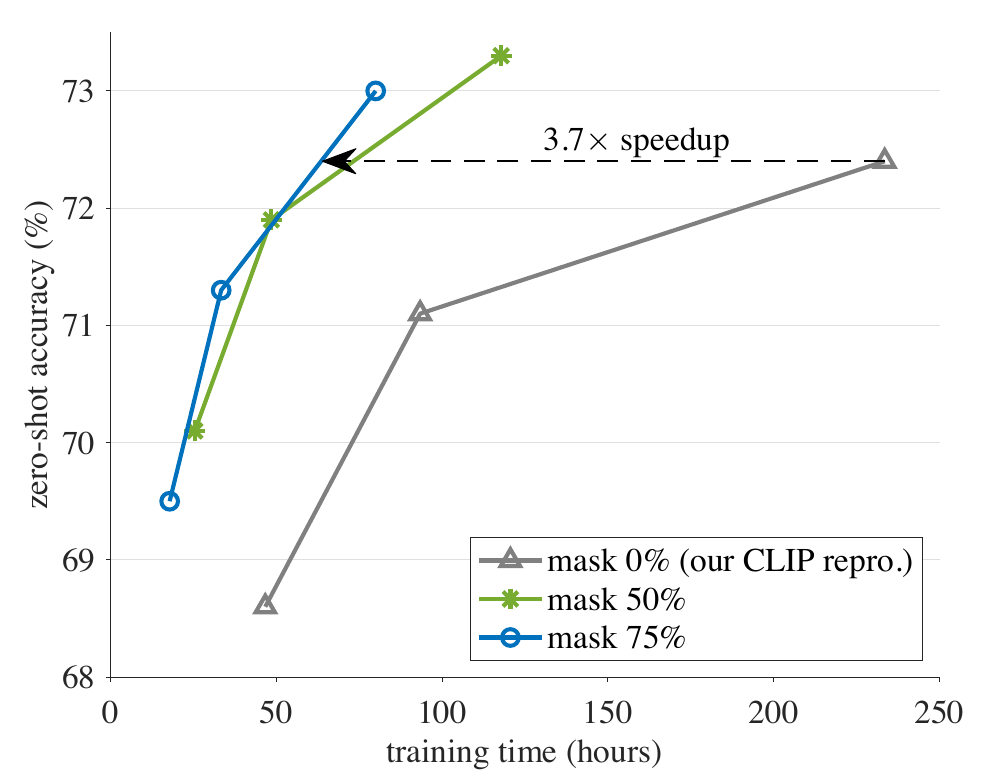}
\vspace{-.5em}
\caption{\textbf{Accuracy \vs training time trade-off}. With a high masking ratio of 50\% or 75\%, our FLIP method trains faster and is more accurate than its CLIP counterpart. 
All entries are benchmarked in 256 TPU-v3 cores. Training is done on LAION-400M for 6.4, 12.8, or 32 epochs, for each masking ratio. Accuracy is evaluated by zero-shot transfer on the ImageNet-1K validation set. The model is ViT-L/16 \cite{Dosovitskiy2021}. More details are in Fig.~\ref{fig:tradeoff_full}. As the CLIP baseline takes $\app$2,500 TPU-days training, a speedup of 3.7$\times$ can save $\app$1,800 TPU-days.
}
\label{fig:teaser}
\vspace{-1em}
\end{figure}

\hspace{-.5em}
Language-supervised visual pre-training, \mbox{\eg, CLIP \cite{Radford2021},} 
has been established as a simple yet powerful methodology for learning representations. Pre-trained CLIP models stand out for their remarkable versatility: they have strong zero-shot transferability  \cite{Radford2021}; they demonstrate unprecedented quality in text-to-image generation (\eg, \cite{Ramesh2022,Rombach2022}); the pre-trained encoder can improve multimodal and even unimodal visual tasks. Like the role played by supervised pre-training a decade ago \cite{Krizhevsky2012}, language-supervised visual pre-training is new fuel empowering various tasks today.

Unlike classical supervised learning with a pre-defined label set, natural language provides \textit{richer} forms of supervision, \eg, on objects, scenes, actions, context, and their relations, at multiple levels of granularity. 
Due to the complex nature of vision plus language, \textit{large-scale} training is essential for the capability of language-supervised models. For example, the original CLIP models \cite{Radford2021} were trained on 400 million data for 32 epochs---which amount to 10,000 ImageNet \cite{Deng2009} epochs, taking thousands of GPU-days \cite{Radford2021,Ilharco2021}.
Even using high-end infrastructures, the wall-clock training time is still a major bottleneck hindering explorations on scaling vision-language learning.

We present Fast Language-Image Pre-training (FLIP), a simple method for efficient CLIP training. Inspired by the sparse computation of Masked Autoencoders (MAE) \cite{He2021}, we randomly remove a large portion of image patches during training. This design introduces a trade-off between ``\textit{how carefully we look at a sample pair}" \vs ``\textit{how many sample pairs we can process}". Using masking, we can: (\textit{i}) see more sample pairs (\ie, more epochs) under the same wall-clock training time, and (\textit{ii}) compare/contrast more sample pairs at each step (\ie, larger batches) under similar memory footprint. Empirically, the benefits of processing more sample pairs greatly outweigh the degradation of per-sample encoding, resulting in a favorable trade-off.

By removing 50\%-75\% patches of a training image, our method reduces computation by 2-4$\x$; it also allows using 2-4$\x$ larger batches with little extra memory cost, which boost accuracy thanks to the behavior of contrastive learning \cite{He2020,Chen2020}.
As summarized in Fig.~\ref{fig:teaser}, FLIP trains $>$3$\x$ faster in wall-clock time for reaching similar accuracy as its CLIP counterpart; with the same number of epochs, FLIP reaches higher accuracy than its CLIP counterpart
while still being 2-3$\x$ faster.

We show that FLIP is a competitive alternative to CLIP on various downstream tasks. 
Pre-trained on the same LAION-400M dataset \cite{Schuhmann2022}, FLIP dominantly outperforms its CLIP counterparts (OpenCLIP \cite{Ilharco2021} and our own reproduction), as evaluated on a large variety of downstream datasets and transfer scenarios. These comparisons suggest that FLIP can readily enjoy the faster training speed while still providing accuracy gains.

Facilitated by faster training, we explore \textit{scaling} FLIP pre-training. We study these three axes: (\textit{i}) scaling model size, (\textit{ii}) scaling dataset size, or (\textit{iii}) scaling training schedule length. We analyze the scaling behavior through carefully controlled experiments. We observe that model scaling and data scaling can both improve accuracy, and data scaling can show gains at no extra training cost.
We hope our method, results, and analysis will encourage future research on scaling vision-language learning.

\section{Related Work}
\label{sec:related}

\paragraph{Learning with masking.} Denoising Autoencoders \cite{Vincent2008} with masking noise \cite{Vincent2010} were proposed as an unsupervised representation learning method over a decade ago. One of its most outstanding applications is masked language modeling represented by BERT \cite{Devlin2019}. In computer vision, explorations along this direction include predicting large missing regions \cite{Pathak2016}, sequence of pixels \cite{Chen2020c}, patches \cite{Dosovitskiy2021,He2021,Xie2021a}, or pre-computed features \cite{Bao2021,Wei2021}.

The Masked Autoencoder (MAE) method \cite{He2021} further takes advantage of masking to reduce training time and memory. MAE sparsely applies the ViT encoder \cite{Dosovitskiy2021} to visible content. It also observes that a high masking ratio is beneficial for accuracy.
The MAE design has been applied to videos \cite{Tong2022,Feichtenhofer2022}, point clouds \cite{Pang2022}, graphs \cite{Tan2022,Chen2022,Hou2022a}, audio \cite{Baade2022,Niizumi2022,Chong2022,Huang2022}, visual control \cite{Xiao2022,Seo2022}, vision-language \cite{Geng2022,Kwon2022,He2022a,Dong2022}, and other modalities \cite{Bachmann2022}.

Our work is related to MAE and its vision-language extensions \cite{Geng2022,Kwon2022,He2022a,Dong2022}. However, our focus is on the \mbox{\textit{scaling}} aspect enabled by the sparse computation; we address the challenge  of \textit{large-scale} CLIP training \cite{Radford2021}, while previous works \cite{Geng2022,Kwon2022,He2022a,Dong2022} are limited in terms of scale. Our method does not perform reconstruction and is not a form of autoencoding. 
Speeding up training by masking is studied in \cite{Wu2022} for self-supervised contrastive learning, \eg, for MoCo \cite{He2020} or BYOL \cite{Grill2020}, but its accuracy could be limited by the scaling behavior of image-only contrastive learning.

\paragraph{Language-supervised learning.} In the past years, CLIP \cite{Radford2021} and related works (\eg, \cite{Jia2021,Pham2021}) have popularized learning visual representations with language supervision. CLIP is a form of contrastive learning \cite{Hadsell2006} by comparing image-text sample pairs. Beyond contrastive learning, generative learning methods have been explored \cite{Desai2021,Wang2021a,Alayrac2022,Yu2022}, optionally combined with contrastive losses \cite{Yu2022}.
Our method focuses on the CLIP method, while we hope it can be extended to generative methods in the future.

\begin{figure}[t]\centering
\includegraphics[width=0.85\linewidth]{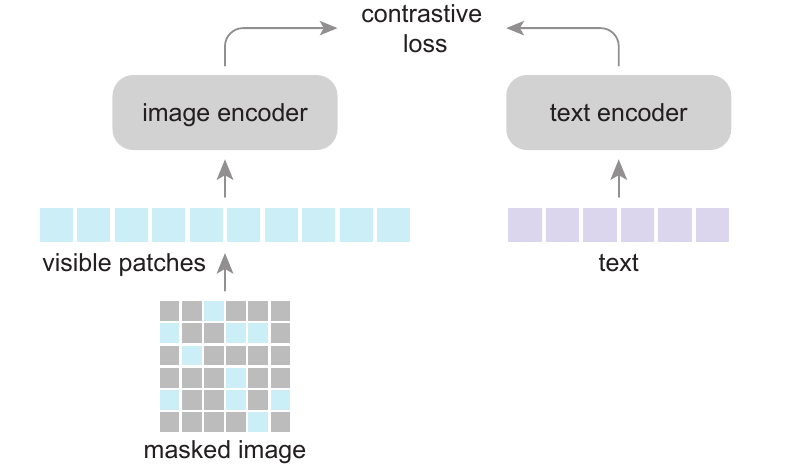}
\caption{\textbf{Our FLIP architecture}. Following CLIP \cite{Radford2021}, we perform contrastive learning on pairs of image and text samples. We randomly mask out image patches with a high masking ratio and encode only the visible patches. We do not perform reconstruction of masked image content. 
}
\label{fig:arch}
\end{figure}

\section{Method}
\label{sec:method}

In a nutshell, our method simply masks out the input data in CLIP \cite{Radford2021} training and reduces computation. See Fig.~\ref{fig:arch}.

In our scenario, the benefit of masking is on wisely spending computation. 
Intuitively, this leads to a trade-off between how densely we encode a sample against how many samples we compare as the learning signal. 
By introducing masking, we can: (\textit{i}) learn from more image-text pairs under the same wall-clock training time, and (\textit{ii}) have a contrastive objective over a larger batch under the same memory constraint. 
We show by experiments that for both aspects, our method is at an advantage in the trade-off.
Next we introduce the key components of our method.

\paragraph{Image masking.} We adopt the Vision Transformer (ViT) \cite{Dosovitskiy2021} as the image encoder. An image is first divided into a grid of non-overlapping patches. We randomly mask out a large portion (\eg, 50\% or 75\%) of patches; the ViT encoder is only applied to the visible patches, following \cite{He2021}.  Using a masking ratio of 50\% (or 75\%) reduces the time complexity of image encoding to 1/2 (or 1/4); it also allows using a 2$\x$ (or 4$\x$) larger batch with a similar memory cost for image encoding.

\paragraph{Text masking.} Optionally, we perform text masking in the same way as image masking. We mask out a portion of the text tokens and apply the encoder only to the visible tokens, as in \cite{He2021}. This is unlike BERT \cite{Devlin2019} that replaces them with a learned mask token. Such sparse computation can reduce the text encoding cost. However, as the text encoder is smaller \cite{Radford2021}, speeding it up does not lead to a better overall trade-off. We study text masking for ablation only.

\paragraph{Objective.} The image/text encoders are trained to minimize a contrastive loss \cite{Oord2018}. The negative sample pairs for contrastive learning consist of other samples in the same batch \cite{Chen2020}.
It has been observed that a large number of negative samples is critical for self-supervised contrastive learning on images \cite{He2020,Chen2020}. This property is more prominent in language-supervised learning.

We do \textit{not} use a reconstruction loss, \mbox{unlike} MAE \cite{He2021}. We find that reconstruction is not necessary for good performance on zero-shot transfer.
Waiving the decoder and reconstruction loss yields a better speedup.

\paragraph{Unmasking.} While the encoder is pre-trained on masked images, it can be directly applied on \textit{intact} images without changes, as is done in \cite{He2021}. This simple setting is sufficient to provide competitive results and will serve as our baseline in ablation experiments.

To close the distribution gap caused by masking, we can set the masking ratio as 0\% and continue pre-training for a small number of steps. This unmasking tuning strategy produces a more favorable accuracy/time trade-off.

\begin{table*}[t]
\vspace{-1em}
\centering
\subfloat[
\textbf{Image masking} yields higher or comparable accuracy and speeds up training. Entries are subject to the same memory limit.
\label{tab:ablation:ratio}
]{
\centering
\begin{minipage}{0.29\linewidth}{\begin{center}
\tablestyle{2pt}{1.1}
\begin{tabular}{z{24}x{24}x{24}x{24}x{24}}
mask & batch & FLOPs & time & acc. \\
\shline
0\% & 16k & 1.00$\x$ & 1.00$\x$ & 68.6 \\
50\% & 32k & 0.52$\x$ & 0.50$\x$ & \textbf{69.6} \\
75\% & 64k & 0.28$\x$ & 0.33$\x$ & 68.2 \\
\end{tabular}
\end{center}}\end{minipage}
}
\hspace{2em}
\subfloat[
\textbf{Batch size}. A large batch has big gains over smaller batches.
\label{tab:ablation:batch}
]{
\begin{minipage}{0.29\linewidth}{\begin{center}
\tablestyle{4pt}{1.1}
\begin{tabular}{x{36}x{36}x{36}}
batch & mask 50\% & mask 75\%\\
\shline
16k & 68.5 & 65.8 \\
32k & 69.6 & 67.3 \\
64k & \textbf{70.4} & \textbf{68.2} \\
\end{tabular}
\end{center}}\end{minipage}
}
\hspace{2em}
\subfloat[
\textbf{Text masking} performs decently, but the speed gain is marginal as its encoder is smaller. Here the image masking ratio is 75\%.
\label{tab:ablation:text}
]{
\begin{minipage}{0.29\linewidth}{\begin{center}
\tablestyle{1pt}{1.1}
\begin{tabular}{z{52}x{36}x{24}x{24}}
\multicolumn{1}{c}{text mask} & text len & time & acc. \\
\shline
baseline, 0\% & 32 & 1.00$\times$ & \textbf{68.2} \\
random, 50\% & 16 & 0.92$\x$ & 66.0 \\
prioritized, 50\% & 16 & 0.92$\x$ & 67.8 \\
\end{tabular}
\end{center}}\end{minipage}
}
\\
\centering
\vspace{1em}
\subfloat[
\textbf{Inference unmasking}. Inference on intact images performs strongly even without tuning.
\label{tab:ablation:infer}
]{
\begin{minipage}{0.29\linewidth}{\begin{center}
\tablestyle{2pt}{1.1}
\begin{tabular}{y{64}x{36}x{36}}
& mask 50\% & mask 75\% \\
\shline
w/ mask & 66.4 & 60.9 \\
w/ mask, ensemble & 68.1 & 65.1 \\
w/o mask & \textbf{69.6} & \textbf{68.2} \\
\end{tabular}
\end{center}}\end{minipage}
}
\hspace{2em}
\subfloat[
\textbf{Unmasked tuning}. The distribution shift by masking is reduced by a short tuning.
\label{tab:ablation:tune}
]{
\centering
\begin{minipage}{0.29\linewidth}{\begin{center}
\tablestyle{4pt}{1.1}
\begin{tabular}{x{36}x{36}x{36}}
& mask 50\% & mask 75\% \\
\shline
baseline & 69.6 & 68.2 \\
+ tuning & \textbf{70.1} & \textbf{69.5} \\
\multicolumn{2}{c}{} \\ %
\end{tabular}
\end{center}}\end{minipage}
}
\hspace{2em}
\subfloat[
\textbf{Reconstruction}. Adding the MAE reconstruction loss has no gain.
\label{tab:ablation:reconstruct}
]{
\begin{minipage}{0.29\linewidth}{\begin{center}
\tablestyle{4pt}{1.1}
\begin{tabular}{x{48}x{36}x{36}}
& mask 50\% & mask 75\% \\
\shline
baseline & \textbf{69.6} & \textbf{68.2} \\
+ MAE & 69.4 & 67.9 \\
\multicolumn{2}{c}{} \\ %
\end{tabular}
\end{center}}\end{minipage}
}
\vspace{-1.5em}
\caption{\textbf{Zero-shot ablation experiments}. Pre-training is on LAION-400M for 6.4 epochs, evaluated by zero-shot classification accuracy on ImageNet-1K validation. The backbone is ViT-L/16 \cite{Dosovitskiy2021}. Unless specified, the baseline setting is: image masking is 50\% (batch 32k) or 75\% (batch 64k), text masking is 0\%, and no unmasked tuning is used.
}
\label{tab:ablations}
\vspace{-.5em}
\end{table*}

\subsection{Implementation}
\label{sec:impl}

Our implementation follows CLIP \cite{Radford2021} and OpenCLIP \cite{Ilharco2021}, with a few modifications we describe in the following. Hyper-parameters are in the appendix. 

Our image encoder follows the ViT paper \cite{Dosovitskiy2021}. 
We do \textit{not} use the extra LayerNorm \cite{Ba2016} after patch embedding, like \cite{Dosovitskiy2021} but unlike \cite{Radford2021}. 
We use global average pooling at the end of the image encoder. The input size is 224.

Our text encoder is a non-autoregressive Transformer \cite{Vaswani2017}, which is easier to adapt to text masking for ablation. We use a WordPiece tokenizer as in \cite{Devlin2019}. We pad or cut the sequences to a fixed length of 32. We note that CLIP in \cite{Radford2021} uses an autoregressive text encoder, a BytePairEncoding tokenizer, and a length of 77. These designs make marginal differences as observed in our initial reproduction.

The outputs of the image encoder and text encoder are projected to the same-dimensional embedding space by a linear layer. The cosine similarities of the embeddings, scaled by a learnable temperature parameter \cite{Radford2021}, are the input to the InfoNCE loss \cite{Oord2018}.

In zero-shot transfer, we follow the prompt engineering in the code of  \cite{Radford2021}.
We use their provided 7 prompt templates for ImageNet zero-shot transfer.

Our implementation is based on JAX \cite{Bradbury2018} with the t5x library \cite{Roberts2022} for large-scale distributed training. Our training is run on TPU v3 infrastructure.

\section{Experiments}

\subsection{Ablations}
\label{sec:ablations}

We first ablate the FLIP design. The image encoder is ViT-L/16 \cite{Dosovitskiy2021}, and the text encoder has a smaller size as per \cite{Radford2021}. We train on LAION-400M \cite{Ilharco2021} and evaluate zero-shot accuracy on ImageNet-1K \cite{Deng2009} validation.

Table~\ref{tab:ablations} shows the ablations trained for 6.4 epochs. Fig.~\ref{fig:teaser} plots the trade-off for up to 32 epochs \cite{Radford2021}. The results are benchmarked in 256 TPU-v3 cores, unless noted.

\paragraph{Masking ratio.} Table~\ref{tab:ablation:ratio} studies the image masking ratios. Here we scale the batch size accordingly (ablated next), so as to roughly maintain the memory footprint.\footnotemark
The 0\% masking entry indicates our CLIP counterpart. Masking 50\% gives 1.2\% higher accuracy than the CLIP baseline, and masking 75\% is on par with the baseline.
Speed-wise, masking 50\% or 75\% takes only 0.50$\times$ or 0.33$\times$ \textit{wall-clock} training time, thanks to the large reduction on FLOPs.

\footnotetext{Directly comparing TPU memory usage can be difficult due to its memory optimizations. We instead validate GPU memory usage using \cite{Ilharco2021}'s reimplementation of FLIP, and find the memory usage is 25.5G, 23.9G and 24.5G  for masking 0\%, 50\% and 75\% on 256 GPUs.}

\paragraph{Batch size.} We ablate the effect of batch size in Table~\ref{tab:ablation:batch}. 
Increasing the batch size consistently improves accuracy.

Notably, even using \textit{the same batch size} of 16k, our 50\% masking entry has a comparable accuracy (68.5\%, Table~\ref{tab:ablation:batch}) with the 0\% masking baseline (68.6\%, Table~\ref{tab:ablation:ratio}). It is possible that the regularization introduced by masking can reduce overfitting, partially counteracting the negative effect of losing information in this setting.
With a higher masking ratio of 75\%, the negative effect is still observed when keeping the batch size unchanged.

Our masking-based method naturally encourages using large batches.
There is little extra memory cost if we scale the batch size according to the masking ratio, as we report in Table~\ref{tab:ablation:ratio}.
In practice, the available memory is always a limitation for larger batches. For example, the setting in Table~\ref{tab:ablation:ratio} has reached the memory limit in our high-end infrastructure (256 TPU-v3 cores with 16GB memory each).\footnotemark~The memory issue is more demanding if using fewer devices, and the gain of our method would be more prominent due to the nearly free increase of batch sizes.

\footnotetext{The ``mask 50\%, 64k'' entry in Table~\ref{tab:ablation:batch} requires 2$\x$ memory \vs those in Table~\ref{tab:ablation:ratio}. This entry can be run using 2$\x$ devices; instead, it can also use memory optimization (\eg, activation checkpointing \cite{Chen2016}) that trades time with memory, which is beyond the focus of this work.}

\paragraph{Text masking.} Table~\ref{tab:ablation:text} studies text masking.
Randomly masking 50\% text decreases accuracy by 2.2\%. 
This is in line with the observation \cite{He2021} that language data has higher information-density than images and thus the text masking ratio should be lower.

As variable-length text sequences are padded to generate a fixed-length batch, we can prioritize masking out the padding tokens. Prioritized sampling preserves more valid tokens than randomly masking the padded sequences uniformly.
It reduces the degradation to 0.4\%.

While our text masking is more aggressive than typical masked language models (\eg, 15\% in \cite{Devlin2019}), the overall speed gain is marginal. This is because the text encoder is smaller and the text sequence is short. The text encoder costs merely 4.4\% computation \vs the image encoder (without masking).
Under this setting, text masking is not a worthwhile trade-off and we do \textit{not} mask text in our other experiments.

\paragraph{Inference unmasking.} By default, we apply our models on intact images at inference-time, similar to \cite{He2021}.
While masking creates a distribution shift between training and inference, simply ignoring this shift works surprisingly well (Table~\ref{tab:ablation:infer}, `w/o mask'), even under the zero-shot setting where no training is ever done on full images.

Table~\ref{tab:ablation:infer} reports that if using masking at inference time, the accuracy drops by a lot (\eg, 7.3\%). This drop can be partially caused by information loss at inference, so we also compare with ensembling multiple masked views \cite{Chen2020c}, where the views are complementary to each other and put together cover all patches. Ensembling reduces the gap (Table~\ref{tab:ablation:infer}), but still lags behind the simple full-view inference.  

\begin{figure}[t]\centering
\vspace{-.5em}
\includegraphics[width=0.98\linewidth]{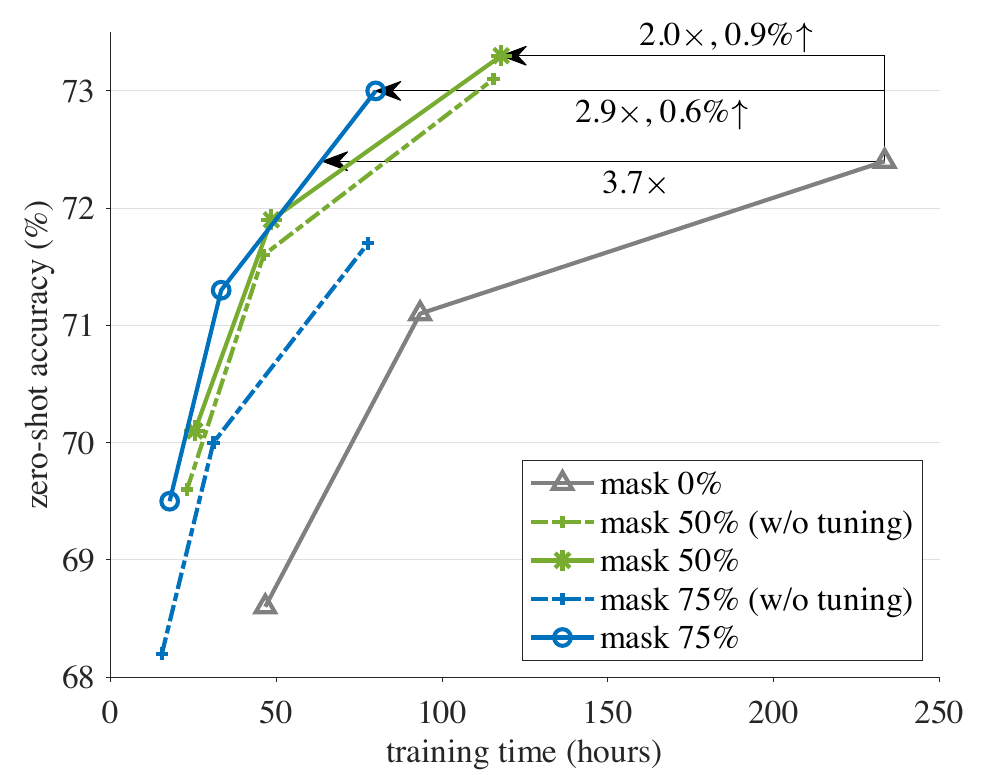}
\vspace{-.5em}
 \caption{\textbf{Accuracy \vs training time trade-off in detail}.
 The setting follows Table~\ref{tab:ablation:ratio}.
Training is for 6.4, 12.8, or 32 epochs, for each masking ratio. Unmasked tuning, if applied, is for 0.32 epoch. All are benchmarked in 256 TPU-v3 cores. Zero-shot accuracy is on IN-1K validation. The model is ViT-L/16.
Our method speeds up training and increases accuracy.
 }
\label{fig:tradeoff_full}
\vspace{-.5em}
\end{figure}

\begin{table*}[t]
\tablestyle{8pt}{1.1}
\begin{tabular}{*l ^c ^c | ^c ^c ^c ^c ^c }
case & data & epochs & B/16 & L/16 & L/14 & H/14 \\
\shline
\rowstyle{\color{dt}}CLIP \cite{Radford2021} & WIT-400M & 32 & 68.6 & - & 75.3 & - \\
OpenCLIP \cite{Ilharco2021} & LAION-400M & 32 & 67.1 & - & 72.8 & - \\
CLIP, our repro. & LAION-400M & 32 & 68.2 & 72.4 & 73.1 & - \\
\hline
\textbf{FLIP} & LAION-400M & 32 & 68.0 & 74.3 & 74.6 & 75.5 \\
\end{tabular}
\vspace{-1em}
\caption{\textbf{Zero-shot accuracy on ImageNet-1K classification}, compared with various CLIP baselines. 
The image size is 224.
The entries noted by {\color{dt}grey} are pre-trained on a different dataset.
Our models use a 64k batch, 50\% masking ratio, and unmasked tuning.
}
\vspace{-.5em}
\label{tab:imagenet:zeroshot}
\end{table*}

\begin{table*}[t]
\tablestyle{6pt}{1.1}
\begin{tabular}{*l ^c ^c ^c | ^c ^c ^c }
case & data & epochs & model & zero-shot & linear probe & fine-tune \\
\shline
\rowstyle{\color{dt}}CLIP \cite{Radford2021} & WIT-400M & 32 & L/14 & 75.3 & 83.9$^\dagger$ & - \\
\rowstyle{\color{dt}}CLIP \cite{Radford2021}, our transfer & WIT-400M & 32 & L/14 & 75.3 & 83.0 & 87.4 \\
OpenCLIP \cite{Ilharco2021} & LAION-400M & 32 & L/14 & 72.8 & 82.1 & 86.2 \\
CLIP, our repro. & LAION-400M & 32 & L/16 & 72.4 & 82.6 & 86.3 \\
\hline
\textbf{FLIP} & LAION-400M & 32 & L/16 & 74.3 & 83.6 & 86.9 \\
\end{tabular}
\vspace{-0.6em}
\caption{\textbf{Linear probing and fine-tuning accuracy on ImageNet-1K classification}, compared with various CLIP baselines.
The entries noted by {\color{dt}grey} are pre-trained on a different dataset.
The image size is 224.
{\footnotesize{$^\dagger$: CLIP in \cite{Radford2021} optimizes with L-BFGS; we use SGD instead.}}
}
\vspace{-1em}
\label{tab:imagenet:learn}
\end{table*}

\paragraph{Unmasked tuning.} Our ablation experiments thus far do \textit{not} involve unmasked tuning. Table~\ref{tab:ablation:tune} reports the results of unmasked tuning for extra 0.32 epoch on the pre-training dataset. It increases accuracy by 1.3\% at the high masking ratio of 75\%, suggesting that tuning can effectively reduce the distribution gap between pre-training and inference.

Fig.~\ref{fig:tradeoff_full} plots the trade-off affected by unmasked tuning (solid \vs dashed).
Unmasked tuning leads to a more desirable trade-off for 75\% masking; it has a comparable trade-off for 50\% masking but improves final accuracy.

\paragraph{Reconstruction.} In Table~\ref{tab:ablation:reconstruct} we investigate adding a reconstruction loss function. The reconstruction head follows the design in MAE \cite{He2021}: it has a small decoder and reconstructs normalized image pixels. The reconstruction loss is added to the contrastive loss.

Table~\ref{tab:ablation:reconstruct} shows that reconstruction has a small negative impact for zero-short results. 
We also see a similar though slight less drop for fine-tuning accuracy on ImageNet.
While this can be a result of suboptimal hyper-parameters (\eg, balancing two losses), to strive for simplicity, we decide \textit{not} to use the reconstruction loss.
Waiving the reconstruction head also helps simplify the system and improves the accuracy/time trade-off.

\paragraph{Accuracy \vs time trade-off.}
Fig.~\ref{fig:tradeoff_full} presents a detailed view on the accuracy \vs training time trade-off. We extend the schedule to up to 32 epochs \cite{Radford2021}.

As shown in Fig.~\ref{fig:tradeoff_full}, FLIP has a clearly better trade-off than the CLIP counterpart. It can achieve similar accuracy as CLIP while enjoying a speedup of $>$3$\times$. With the same 32-epoch schedule, our method is \app1\% more accurate than the CLIP counterpart and 2$\times$ faster (masking 50\%).

The speedup of our method is of great practical value. The CLIP baseline takes \app10 days training in 256 \mbox{TPU-v3} cores, so a speedup of 2-3$\times$ saves many days in wall-clock time. This speedup facilitates exploring the scaling behavior, as we will discuss later in Sec.~\ref{sec:scaling}.

\subsection{Comparisons with CLIP}

In this section, we compare with various CLIP baselines in a large variety of scenarios. We show that our method is a competitive alternative to CLIP; as such, our fast training method is a more desirable choice in practice.

\newcolumntype{S}{@{}>{\lrbox0}l<{\endlrbox}}  %
\definecolor{lightgreen}{HTML}{D8ECD1}
\newcommand{\better}[1]{\colorbox{lightgreen}{#1}}
\newcommand{\datatag}[1]{\rotatebox[origin=l]{90}{\scriptsize{#1}}}
\begin{table*}
\vspace{-1.em}
\centering
\resizebox{1.0\linewidth}{!}{
\tablestyle{1.5pt}{1.1}

\begin{tabular}{*l S ^c | ^c ^c ^c ^c ^c ^c ^c ^c ^c ^c ^c ^c ^c ^c ^c ^c ^c ^c ^c ^c ^c ^c ^c ^c ^c ^c}
& model & data  & \datatag{Food101} & \datatag{CIFAR10} & \datatag{CIFAR100} & \datatag{Birdsnap} & \datatag{SUN397} & \datatag{Cars} & \datatag{Aircraft} & \datatag{VOC2007} & \datatag{DTD} & \datatag{Oxford Pets} & \datatag{Caltech101} & \datatag{Flowers102} & \datatag{MNIST} & \datatag{STL10} & \datatag{EuroSAT} & \datatag{RESISC45} & \datatag{GTSRB} & \datatag{KITTI} & \datatag{Country211} & \datatag{PCam} & \datatag{UCF101} & \datatag{Kinetics700} & \datatag{CLEVR} & \datatag{HatefulMemes} & \datatag{SST2}  \\
\shline

\rowstyle{\color{dt}}CLIP \cite{Radford2021} & L/14 & \scriptsize WIT-400M & 92.9 & 96.2 & 77.9 & 48.3 & 67.7 & 77.3 & 36.1 & 84.1 & 55.3 & 93.5 & 92.6 & 78.7 & 87.2 &    99.3 & 59.9 & 71.6 & 50.3 & 23.1 & 32.7 & 58.8 & 76.2 & 60.3 & 24.3 & 63.3 & 64.0 \\

\rowstyle{\color{dt}}CLIP  \cite{Radford2021}, our eval. & L/14 & \scriptsize WIT-400M & 
91.0 & 95.2 & 75.6 & 51.2 & 66.6 & 75.0 & 32.3 & 83.3  & 55.0 & 93.6 & 92.4 & 77.7 & 76.0 &    99.3 & 62.0 & 71.6 & 51.6 & 26.9 & 30.9 & 51.6 & 76.1 & 59.5 & 22.2 & 55.3 & 67.3  \\

OpenCLIP \cite{Ilharco2021}, our eval. & L/14 & \scriptsize LAION-400M & 87.4 & 94.1 & 77.1 & 61.3 & 70.7 & 86.2 & 21.8 & \better{83.5} & 54.9 & 90.8 & \better{94.0} & 72.1 & 71.5 &    98.2 & 53.3 & 67.7 & 47.3 & 29.3 & 21.6 & 51.1 & 71.3 & 50.5 & 22.0 & \better{55.3} & 57.1 \\

CLIP, our repro. & L/14 & \scriptsize LAION-400M &  88.1 & 96.0 & 81.3 & 60.5 & 72.3 & 89.1 & 25.8 & 81.1 & 59.3 & \better{93.2} & 93.2 & 74.6 & 69.1 &   96.5 & 50.7 & 69.2 & \better{50.2} &  29.4 & 21.4 & \better{53.1} & 71.5 & 53.5 & 18.5 & 53.3 & 57.2   \\

\hline

\textbf{FLIP} & L/14 & \scriptsize LAION-400M & \better{89.3} & \better{97.2} & \better{84.1} & \better{63.0} & \better{73.1} & \better{90.7} & \better{29.1} & 83.1 & \better{60.4} & 92.6 & 93.8 & \better{75.0} & \better{80.3} &   \better{98.5} & \better{53.5} & \better{70.8} & 41.4 & \better{34.8} & \better{23.1} & 50.3 & \better{74.1} & \better{55.8} & \better{22.7} & 54.0 & \better{58.5}  \\

\end{tabular}
}
\vspace{-.5em}
\caption{\textbf{Zero-shot accuracy on more classification datasets}, compared with various CLIP baselines. 
This table follows Table 11 in \cite{Radford2021}. The model is ViT-L/14 with an image size of 224, for all entries. Entries in green are the best ones using the LAION-400M data.}

\label{tab:more:zeroshot}
\vspace{1em}
\tablestyle{3pt}{1.05}
\begin{tabular}{*l ^l ^c | ^c ^c ^c | ^c ^c ^c | ^c ^c ^c | ^c ^c ^c}
& & & \multicolumn{6}{c|}{\scriptsize text retrieval} & \multicolumn{6}{c}{\scriptsize image retrieval} \\
& & & \multicolumn{3}{c|}{\scriptsize Flickr30k} & \multicolumn{3}{c|}{\scriptsize COCO} & \multicolumn{3}{c|}{\scriptsize Flickr30k} & \multicolumn{3}{c}{\scriptsize COCO} \\
case & model & data & \scriptsize R@1 & \scriptsize R@5 & \scriptsize R@10 & \scriptsize R@1 & \scriptsize R@5 & \scriptsize R@10 & \scriptsize R@1 & \scriptsize R@5 & \scriptsize R@10 & \scriptsize R@1 & \scriptsize R@5 & \scriptsize R@10 \\
\shline

\rowstyle{\color{dt}}CLIP \cite{Radford2021} & L/14@336 & \scriptsize WIT-400M & 88.0 & 98.7 & 99.4 & 58.4 & 81.5 & 88.1 & 68.7 & 90.6 & 95.2 & 37.8 & 62.4 & 72.2\\

\rowstyle{\color{dt}}CLIP \cite{Radford2021}, our eval. & L/14@336 & \scriptsize WIT-400M & 88.9 & 98.7 & 99.9 & 58.7 & 80.4	 & 87.9 & 72.5 & 91.7 & 95.2 & 38.5 & 62.8 & 72.5 \\

\rowstyle{\color{dt}}CLIP \cite{Radford2021}, our eval. & L/14 & \scriptsize WIT-400M & 87.8 & 99.1 & 99.8 & 56.2 & 79.8 & 86.4 & 69.3 & 90.2 & 94.0 & 35.8 & 60.7 & 70.7\\

OpenCLIP \cite{Ilharco2021}, our eval. & L/14 & \scriptsize LAION-400M & 87.3 & 97.9 & 99.1 & 58.0 & 80.6 & 88.1 & 72.0 & 90.8 & 95.0 & 41.3 & 66.6 & 76.1 \\

CLIP, our impl. & L/14 & \scriptsize LAION-400M & 87.4 & 98.4 & 99.5 & 59.1 & 82.5 & 89.4 & 74.4 & 92.2 & 95.5 & 43.2 & 68.5 & 77.5 \\

\hline

FLIP & L/14 & \scriptsize LAION-400M & \better{89.1} & \better{98.5} & \better{99.6} & \better{60.2} & \better{82.6} & \better{89.9} & \better{75.4} & \better{92.5} & \better{95.9} & \better{44.2} & \better{69.2} & \better{78.4} \\

\end{tabular}
\vspace{-.5em}
\caption{\textbf{Zero-shot image/text retrieval}, compared with various CLIP baselines. 
The image size is 224 if not noted.
Entries in green are the best ones using the LAION-400M data.
}
\label{tab:retrieval}
\vspace{0.em}
\tablestyle{4pt}{1.05}
\begin{tabular}{*l ^l ^c | ^c ^c ^c ^c ^c ^c ^c ^c ^c }
 & & & \scriptsize IN-V2 & \scriptsize IN-A & \scriptsize IN-R & \scriptsize ObjectNet & \scriptsize IN-Sketch & \multicolumn{2}{c}{\scriptsize IN-Vid} & \multicolumn{2}{c}{\scriptsize YTBB} \\
  & model  & data   & \scriptsize top-1 & \scriptsize top-1 & \scriptsize top-1 & \scriptsize top-1 & \scriptsize top-1 & \scriptsize PM-0 & \scriptsize PM-10 & \scriptsize PM-0 & \scriptsize PM-10 \\
\shline
\rowstyle{\color{dt}}CLIP \cite{Radford2021} & L/14@336 & \scriptsize WIT-400M  & 70.1 & 77.2 & 88.9 & 72.3 & 60.2 & 95.3 & 89.2 & 95.2 & 88.5 \\
\rowstyle{\color{dt}}CLIP \cite{Radford2021}, our eval. & L/14@336 & \scriptsize WIT-400M & 70.4 & 78.0 & 89.0 & 69.3 & 59.7 & 95.9 & 88.8 & 95.3 & 89.4 \\
\rowstyle{\color{dt}}CLIP \cite{Radford2021}, our eval. & L/14 & \scriptsize WIT-400M  & 69.5 & 71.9 & 86.8 & 68.6 & 58.5 & 94.6 & 87.0 & 94.1 & 86.4 \\

OpenCLIP \cite{Ilharco2021}, our eval. & L/14 & \scriptsize LAION-400M & 64.0 & 48.3 & 84.3 & 58.8 & 56.9 & 90.3 & 81.4 & 86.5 & 77.8  \\
CLIP, our repro. & L/14 & \scriptsize LAION-400M  & 65.6 & 46.3 & 84.7 & 58.0 & 58.7 & 89.3 & 80.5 & 85.7 & 77.8 \\
\hline
\textbf{FLIP} & L/14 & \scriptsize LAION-400M  & \better{66.8} & \better{51.2} & \better{86.5} & \better{59.1} &\better{59.9} & \better{91.1} & \better{83.5} & \better{89.4} & \better{83.3}
\end{tabular}
\vspace{-.5em}
\caption{\textbf{Zero-shot robustness evaluation,} compared with various CLIP baselines. This table follows Table 16 in \cite{Radford2021}.
The image size is 224 if not noted. Entries in green are the best ones using the LAION-400M data.
}
\label{tab:imagenet:robust}
\vspace{1em}
\tablestyle{5pt}{1.1}
\begin{tabular}{*l ^l ^c | ^c ^c ^c ^c ^c | ^c ^c | ^c ^c ^c | ^c ^c ^c}
& & & \multicolumn{5}{c|}{COCO caption} & \multicolumn{2}{c|}{nocaps} & \multicolumn{1}{c}{VQAv2} \\
case & model & data & \scriptsize BLEU-4 & \scriptsize METEOR & \scriptsize ROUGE-L & \scriptsize CIDEr & \scriptsize SPICE & \scriptsize CIDEr & \scriptsize SPICE & acc. \\\shline
\rowstyle{\color{dt}}CLIP \cite{Radford2021}, our transfer & L/14 & WIT-400M &
37.5 & 29.6 & 58.7 & 126.9 & 22.8 & 82.5 & 12.1 & 76.6 \\
OpenCLIP \cite{Ilharco2021}, our transfer & L/14 & LAION-400M &
36.7 & 29.3 & 58.4 & 125.0 & 22.7 & 83.4 & 12.3 & 74.5 \\
CLIP, our repro. & L/16 & LAION-400M &
36.4 & 29.3 & 58.4 & 125.6 & 22.8 & 82.8 & 12.2 & 74.5 \\
\hline
FLIP & L/16 & LAION-400M &
\better{37.4} & \better{29.5} & \better{58.8} & \better{127.7} & \better{23.0} & \better{85.9} & \better{12.4} & \better{74.7} \\
\end{tabular}
\vspace{-.5em}
\caption{\textbf{Image Captioning and Visual Question Answering}, compared with various CLIP baselines. 
Entries in green are the best ones using LAION-400M. Here the results are on the COCO captioning test split of \cite{Karpathy2015}, nocaps val split, and VQAv2 test-dev split, respectively.}
\vspace{-.5em}
\label{tab:captioning_and_vqa}
\end{table*}

\vspace{.5em}
We consider the following CLIP baselines:
\begin{itemize}
\setlength\itemsep{.1em}
	\item The original CLIP checkpoints \cite{Radford2021}, trained on the private dataset WIT-400M.
	\item OpenCLIP \cite{Ilharco2021}, trained on LAION-400M.
	\item Our CLIP reproduction, trained on LAION-400M.
\end{itemize}

\noindent
The original CLIP \cite{Radford2021} was trained on a private dataset, so a direct comparison with it should reflect the effect of data, not just methods.
OpenCLIP \cite{Ilharco2021} is a faithful reproduction of CLIP yet trained on a public dataset that we can use, so it is a good reference for us to isolate the effect of dataset differences. Our CLIP reproduction further helps isolate other implementation subtleties and allows us to pinpoint the effect of the FLIP method.

For all tasks studied in this subsection, we compare with all these CLIP baselines. This allows us to better understand the influence of the data and of the methods.

\paragraph{ImageNet zero-shot transfer.} In Table~\ref{tab:imagenet:zeroshot} we compare with the CLIP baselines on ImageNet-1K \cite{Deng2009} zero-shot transfer.

As a sanity check, our CLIP reproduction has slightly higher accuracy than OpenCLIP trained on the same data.
The original CLIP has higher accuracy than our reproduction and OpenCLIP, which could be caused by the difference between the pre-training datasets.

Table~\ref{tab:imagenet:zeroshot} reports the results of our FLIP models, using the best practice as we have ablated in Table~\ref{tab:ablations} (a 64k batch, 50\% masking ratio, and unmasked tuning).
For \mbox{ViT-L/14},\footnotemark~our method has 74.6\% accuracy, which is 1.8\% higher than OpenCLIP and 1.5\% higher than our CLIP reproduction. Comparing with the original CLIP, our method reduces the gap to 0.7\%. We hope our method will improve the original CLIP result if it were trained on the WIT data.

\footnotetext{For a legacy reason, we pre-trained our ViT-L models with a patch size of 16, following the original ViT paper \cite{Dosovitskiy2021}. The CLIP paper \cite{Radford2021} uses L/14 instead. 
To save resources, we report our L/14 results by tuning the L/16 pre-trained model, in a way similar to unmasked tuning.}

\paragraph{ImageNet linear probing.}
Table~\ref{tab:imagenet:learn} compares the linear probing results, \ie, training a linear classifier on the target dataset with frozen features.
FLIP has 83.6\% accuracy, 1.0\% higher than our CLIP counterpart. It is also 0.6\% higher than our transfer of the original CLIP checkpoint, using the same SGD trainer.

\paragraph{ImageNet fine-tuning.} Table~\ref{tab:imagenet:learn} also compares full fine-tuning results.
Our fine-tuning implementation follows MAE \cite{He2021}, with the learning rate tuned for each entry.
It is worth noting that with our fine-tuning recipe, the \textit{original} CLIP checkpoint reaches 87.4\%, much higher than previous reports \cite{Wortsman2022,Wei2022,Hou2022} on this metric. 
CLIP is still a strong model under the fine-tuning protocol.

FLIP outperforms the CLIP counterparts pre-trained on the same data. 
Our result of 86.9\% (or 87.1\% using L/14) is behind but close to the result of the original CLIP checkpoint's 87.4\%, using our fine-tuning recipe.

\paragraph{Zero-shot classification on more datasets.}
In Table~\ref{tab:more:zeroshot} we compare on the extra datasets studied in \cite{Radford2021}. As the results can be sensitive to evaluation implementation (\eg, text prompts, image pre-processing), we provide our evaluations of the original CLIP checkpoint and OpenCLIP.

Notably, we observe \textit{clear systematic gaps caused by pre-training data}, as benchmarked using the same evaluation code.
The WIT dataset is beneficial for some tasks (\eg, Aircraft, Country211, SST2), while LAION is beneficial for some others (\eg, Birdsnap, SUN397, Cars).

After isolating the influence of pre-training data, we observe that FLIP is \textit{dominantly} better than OpenCLIP and our CLIP reproduction, as marked by \colorbox{lightgreen}{green} in Table~\ref{tab:more:zeroshot}.

\paragraph{Zero-shot retrieval.} Table~\ref{tab:retrieval} reports image/text retrieval results on Flickr30k \cite{Young2014} and COCO \cite{Lin2014}. FLIP outperforms all CLIP competitors, including the original CLIP (evaluated on the same 224 size). The WIT dataset has no advantage over LAION for these two retrieval datasets.

\paragraph{Zero-shot robustness evaluation.} In Table~\ref{tab:imagenet:robust} we compare on robustness evaluation, following \cite{Radford2021}.
We again observe \textit{clear systematic gaps caused by pre-training data}. 
Using the same evaluation code (``our eval'' in Table~\ref{tab:imagenet:robust}), CLIP pre-trained on WIT is clearly better than other entries pre-trained on LAION. 
Taking IN-Adversarial (IN-A) as an example: the LAION-based OpenCLIP \cite{Ilharco2021} has only 48.3\% accuracy (or 46.6\% reported by \cite{Ilharco2021}). While FLIP (51.2\%) can outperform the LAION-based CLIP by a large margin, it is still 20\% below the WIT-based CLIP (71.9\%).

Discounting the influence of pre-training data, our FLIP training has clearly better robustness than its CLIP counterparts in \textit{all cases}. We hypothesize that masking as a form of noise and regularization can improve robustness.

\begin{figure*}[t]\centering
\vspace{-0.5em}
\newcommand{\hei}{0.2\linewidth}
\subfloat[\textbf{Model scaling}\label{fig:scaling:model}]
{\includegraphics[height=\hei]{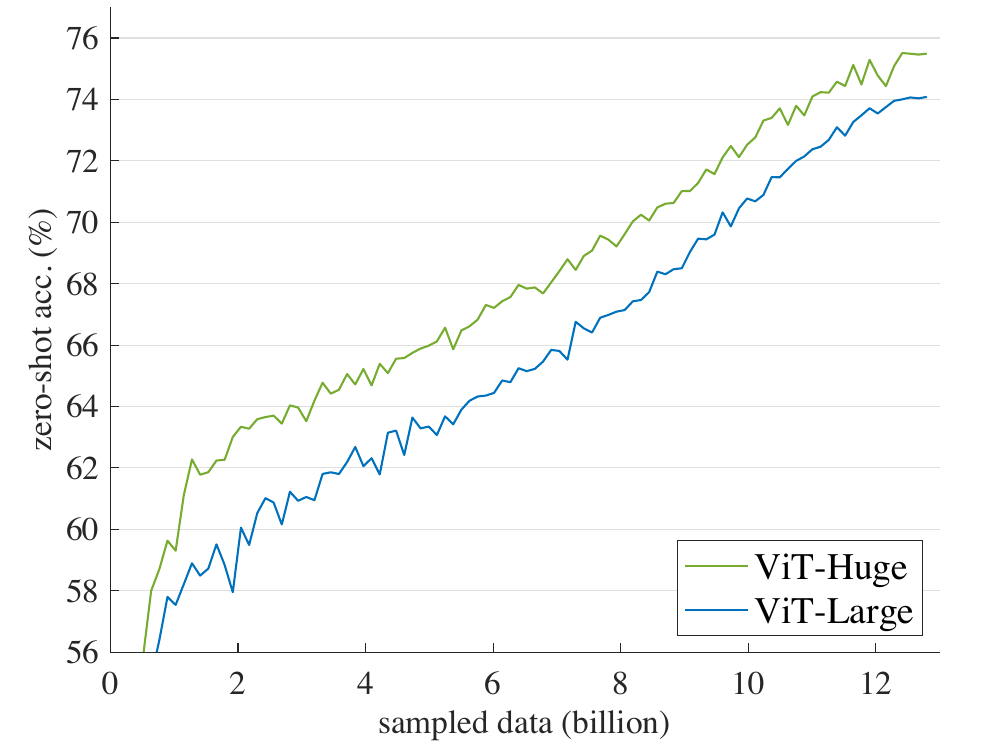}}
\subfloat[\textbf{Data scaling}\label{fig:scaling:data}]
{\includegraphics[height=\hei]{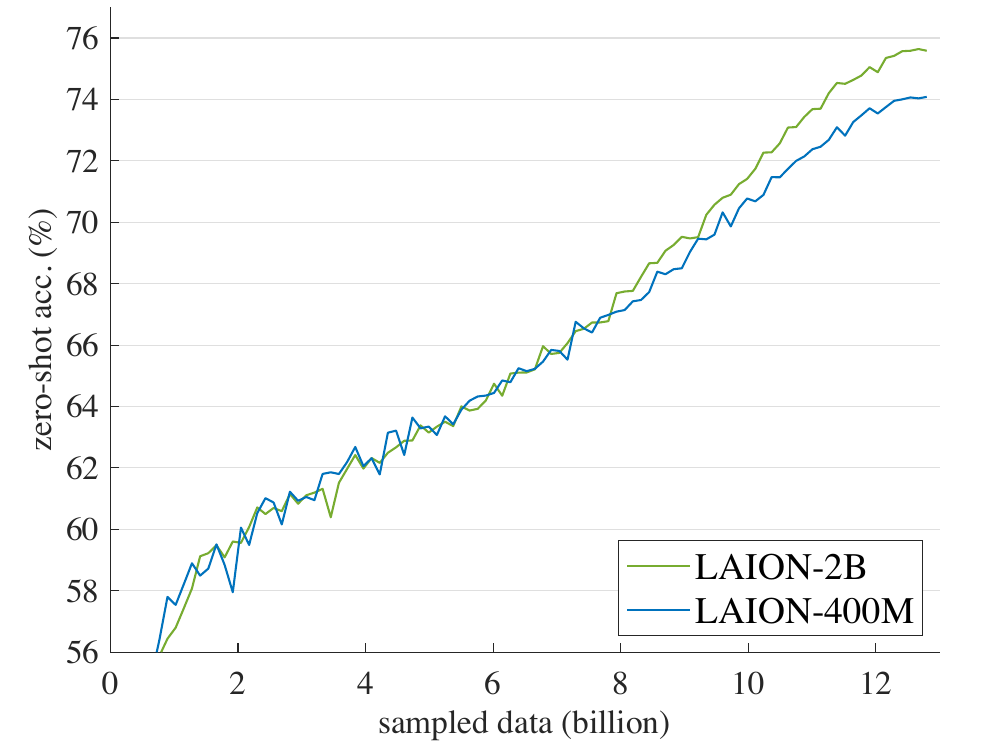}}
\subfloat[\textbf{Schedule scaling} \label{fig:scaling:length}]
{\includegraphics[height=\hei]{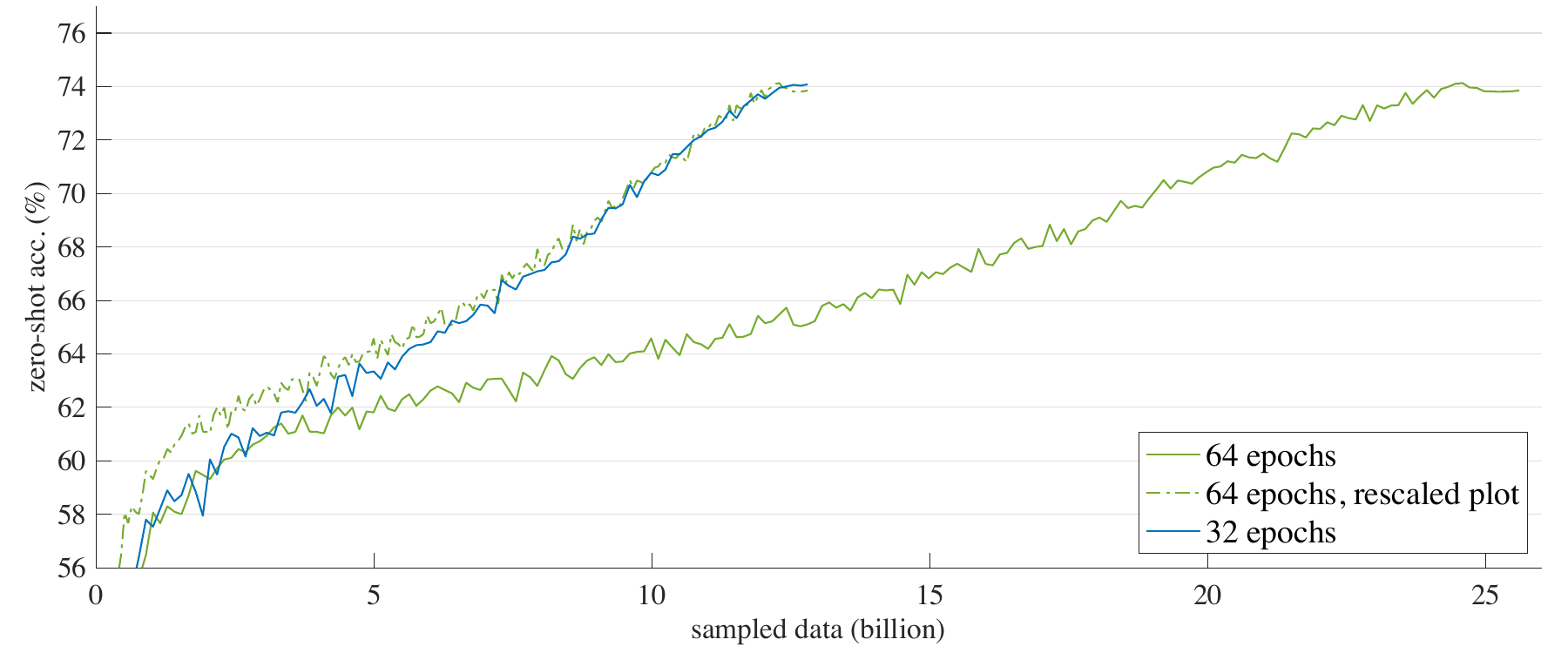}}
\caption{\textbf{Training curves of scaling}. The x-axis is the number of sampled data during training, and the y-axis is the zero-shot accuracy on IN-1K.
The blue curve is the baseline setting: ViT-Large, LAION-400M data, 32 epochs (12.8B sampled data). In each subplot, we compare with scaling one factor on the baseline. In schedule scaling (Fig.~\ref{fig:scaling:length}), 
we plot an extra hypothetical curve for a better visualization. }
\label{fig:scaling}
\end{figure*}

\paragraph{Image Captioning.} See Table~\ref{tab:captioning_and_vqa} for the captioning performance on COCO \cite{Lin2014} and nocaps \cite{Agrawal2019}.
Our captioning implementation follows the cross-entropy training baseline in \cite{Barraco2022}. Unlike classification in which only a classifier layer is added after pre-training, here the fine-tuning model has a newly initialized captioner (detailed in appendix).
In this task, FLIP outperforms the original CLIP checkpoint in several metrics. Compared to our CLIP baseline, which is pre-trained on the same data, FLIP also shows a clear gain, especially in BLEU-4 and CIDEr metrics. 

\paragraph{Visual Question Answering.} We evaluate on the VQAv2 dataset \cite{Goyal2017a}, with a fine-tuning setup following \cite{Dou2022}.
We use a newly initialized multimodal fusion transformer and an answer classifier to obtain the VQA outputs (detailed in appendix).
Table~\ref{tab:captioning_and_vqa} (rightmost column) reports results on VQAv2. 
All entries pre-trained on LAION perform similarly, and CLIP pre-trained on WIT is the best.

\paragraph{{Summary of Comparisons}.} Across a large variety of scenarios, FLIP is \textit{dominantly} better than its CLIP counterparts (OpenCLIP and our reproduction) pre-trained on the same LAION data, in some cases by large margins.

The difference between the WIT data and LAION data can create large systematic gaps, as observed in many downstream tasks. We hope our study will draw attention to these data-dependent gaps in future research.

\begin{table*}[t]
\tablestyle{4pt}{1.1}
\begin{tabular}{*l ^l ^c  ^c | ^c ^c ^c ^c ^c | ^c ^c ^c ^c ^c}
& & & &
\multicolumn{5}{c|}{\textbf{zero-shot transfer}} &
\multicolumn{5}{c}{\textbf{transfer learning}} \\
& & & &
\multicolumn{1}{c|}{\scriptsize zero-shot} & 
\multicolumn{2}{c|}{\scriptsize text retrieval} &
\multicolumn{2}{c|}{\scriptsize image retrieval} &
\scriptsize lin-probe & \multicolumn{1}{c|}{\scriptsize fine-tune} & 
\multicolumn{2}{c|}{\scriptsize captioning} &
\scriptsize vqa \\
case & model & data & sampled &
\multicolumn{1}{c|}{\scriptsize IN-1K} & 
\scriptsize Flickr30k & \multicolumn{1}{c|}{\scriptsize COCO} & 
\scriptsize Flickr30k & \multicolumn{1}{c|}{\scriptsize COCO} & 
\scriptsize IN-1K & \multicolumn{1}{c|}{\scriptsize IN-1K} &
\scriptsize COCO & \multicolumn{1}{c|}{\scriptsize nocaps} &
\scriptsize VQAv2 \\
\shline
baseline & Large & 400M & 12.8B &
74.3 & 
88.4 & 59.8 &
75.0 & 44.1 &
83.6 & 86.9 &
127.7 & 85.9 & 
74.7 \\
\hline
{model scaling} & \textbf{Huge} & 400M & 12.8B & 
75.5 &
89.2 & 62.8 &
76.4 & 46.0 &
\better{84.3} & \better{87.3} &
\better{130.3} & \better{91.5} & 
\better{76.3} \\
{data scaling} & Large & \textbf{2B} & 12.8B &
\better{75.8} &
\better{91.7} & \better{63.8} &
\better{78.2} & \better{47.3} &
84.2 & 87.1 & 
128.9 & 87.0 & 
75.5 \\
{schedule scaling} & Large & 400M & \textbf{25.6B} &
73.9 &
89.7 & 60.1 &
75.5 & 44.4 &
83.7 & 86.9 &
127.9 & 86.8 &
75.0 \\
\hline
{model+data scaling} & \textbf{Huge} & \textbf{2B} & 12.8B &
{77.6} &
{92.8} & {67.0} &
{79.9} & {49.5} &
{85.1} & {87.7} &
\textbf{130.4} & \textbf{92.6} &
{77.1} \\
{joint scaling} & \textbf{Huge} & \textbf{2B} & \textbf{25.6B} &
\textbf{78.8} &
\textbf{93.1} & \textbf{67.8} & 
\textbf{80.9} & \textbf{50.5} &
\textbf{85.6} & \textbf{87.9} &
130.2 & 91.2 &
\textbf{77.3} \\
\end{tabular}
\vspace{-.5em}
\caption{\textbf{Scaling behavior of FLIP}, evaluated on a diverse set of downstream tasks: classification, retrieval (R@1), captioning (CIDEr), and visual question answering. 
In the middle three rows, we scale along one of the three axes (model, data, schedule), and the {green} entries denote the best ones among these three scaling cases. 
Data scaling is in general favored under the zero-shot transfer scenario, while model scaling is in general favored under the transfer learning scenario (\ie, with trainable weights in downstream).
}
\label{tab:scaling}
\vspace{-.5em}
\end{table*}

\subsection{Scaling Behavior}  \label{sec:scaling}

Facilitated by the speed-up of FLIP, we explore the scaling behavior beyond the largest case studied in CLIP \cite{Radford2021}.
We study scaling along either of these three axes:
\begin{itemize}
\setlength\itemsep{.1em}
	\item \textbf{Model scaling}. We replace the \mbox{ViT-L} image encoder with \mbox{ViT-H}, which has $\app$2$\x$ parameters. The text encoder is also scaled accordingly.
	\item \textbf{Data scaling}. We scale the pre-training data from 400 million to 2 billion, using the LAION-2B set \cite{Ilharco2021}. To better separate the influence of more data from the influence of longer training, \textit{we fix the total number of sampled data} (12.8B, which amounts to 32 epochs of 400M data and 6.4 epochs of 2B data).
	\item \textbf{Schedule scaling.} We increase the sampled data from 12.8B to 25.6B (64 epochs of 400M  data).
\end{itemize}

\noindent We study scaling along one of these three axes at each time while keeping others unchanged.
The results are summarized in Fig.~\ref{fig:scaling} and Table~\ref{tab:scaling}.

\paragraph{Training curves.} The three scaling strategies exhibit different trends in training curves (Fig.~\ref{fig:scaling}).

\textit{Model scaling} (Fig.~\ref{fig:scaling:model}) presents a clear gap that persists throughout training, though the gap is smaller at the end.

\textit{Data scaling} (Fig.~\ref{fig:scaling:data}), on the other hand, performs similarly at the first half of training, but starts to present a good gain later. Note that there is \textit{no extra computational cost} in this setting, as we control the total number of sampled data. %

\textit{Schedule scaling} (Fig.~\ref{fig:scaling:length}) trains 2$\times$ longer. To provide a more intuitive comparison, we plot a hypothetical curve that is rescaled by 1/2 along the x-axis (dashed line).
Despite the longer training, the gain is diminishing or none (more numbers in Table~\ref{tab:scaling}).

\paragraph{Transferability.} Table~\ref{tab:scaling} provides an all-around comparison on various downstream tasks regarding the scaling behavior.
Overall, model scaling and data scaling both can consistently outperform the baseline in all metrics, in some cases by large margins. 

We categorize the downstream tasks into two scenarios: (\textit{i}) zero-shot transfer, \ie, no learning is performed on the downstream dataset; (\textit{ii}) transfer learning, \ie, part or all of the weights are trained on the downstream dataset. For the tasks studied here, \textit{data scaling is in general favored for zero-shot transfer, while model scaling is in general favored for transfer learning}. However, it is worth noting that the transfer learning performance depends on the size of the downstream dataset, and training a big model on a too small downstream set is still subject to the overfitting risk.

It is encouraging to see that \textit{data scaling} is clearly beneficial, even \textit{not} incurring longer training \textit{nor} additional computation.
On the contrary, even spending more computation by schedule scaling gives diminishing returns. These comparisons suggest that {large-scale data} are beneficial mainly because they provide richer information.

Next we scale \textit{both} model and data (Table~\ref{tab:scaling}, second last row). For all metrics, model+data scaling improves over scaling either alone. The gains of model scaling and data scaling are highly complementary: \eg, in zero-shot IN-1K, model scaling alone improves over the baseline by 1.2\% (74.3\%$\rightarrow$75.5\%), and data scaling alone improves by 1.5\% (74.3\%$\rightarrow$75.8\%). Scaling both improves by 3.3\% (77.6\%), more than the two deltas combined. This behavior is also observed in several other tasks. This indicates that a larger model desires more data to unleash its potential.

Finally, we report \textit{joint scaling} all three axes (Table~\ref{tab:scaling}, last row). 
Our results show that combining schedule scaling leads to improved performances across most metrics. This suggests that schedule scaling is particularly beneficial when coupled with larger models and larger-scale data.

Our result of 78.8\% on zero-shot IN-1K outperforms the state-of-the-art result trained on public data with ViT-H (78.0\% of \mbox{OpenCLIP}). Also based on \mbox{LAION-2B}, their result is trained with 32B sampled data, 1.25$\times$ more than ours. Given the 50\% masking we use, our training is estimated to be 2.5$\times$ faster than theirs if both were run on the same hardware. As OpenCLIP's result reports a training cost of $\app$5,600 GPU-days, our method could save $\app$3,360 GPU-days based on a rough estimation. Additionally, without enabling 2$\times$ schedule, our entry of ``model+data scaling" is estimated 5$\times$ faster than theirs and can save $\app$4,480 GPU-days. This is considerable cost reduction.

\section{Discussion and Conclusion}

Language is a stronger form of supervision than classical closed-set labels. 
Language provides rich information for supervision. Therefore, \textit{scaling}, which can involve increasing capacity (model scaling) and increasing information (data scaling), is essential for attaining good results in language-supervised training.

CLIP \cite{Radford2021} is an outstanding example of ``\textit{simple algorithms that scale well}". The simple design of CLIP allows it to be relatively easily executed at substantially larger scales and achieve big leaps compared to preceding methods. Our method largely maintains the simplicity of CLIP while pushing it further along the scaling aspect.

Our method can provide a 2-3$\times$ speedup or more. For the scale concerned in this study, such a speedup can reduce  wall-clock time by a great amount (\eg, at the order of thousands of TPU/GPU-days). Besides accelerating research cycles, the speedup can also save a great amount of energy and commercial cost. These are all ingredients of great importance in large-scale machine learning research.

Our study involves controlled comparisons with various CLIP baselines, which help us to break down the gaps contributed by different factors. We show that FLIP outperforms its CLIP counterparts pre-trained on the same LAION data. By comparing several LAION-based models and the original WIT-based ones, we observe that the pre-training data creates big systematic gaps in several tasks.

Our study provides controlled experiments on scaling behavior. 
We observe that data scaling is a favored scaling dimension, given that it can improve accuracy with no extra cost at training or inference time. 
Our fast method encourages us to scale beyond what is studied in this work.

\paragraph{Broader impacts.} Training large-scale models costs high energy consumption and carbon emissions. While our method has reduced such cost to 1/2-1/3, the remaining cost is still sizable. We hope our work will attract more attention to the research direction on reducing the cost of training vision-language models.

The numerical results in this paper are based on a publicly available large-scale dataset \cite{Schuhmann2022}. The resulting model weights will reflect the data biases, including potentially negative implications. When compared using the same data, the statistical differences between two methods should reflect the method properties to a good extent; however, when compared entries using different training data, the biases of the data should always be part of the considerations.

\appendix

\newcommand{\lr}{\emph{lr}\xspace}
\newcommand{\wtd}{\emph{wd}\xspace}
\newcommand{\drp}{\emph{dp}\xspace}
\newcommand{\expnum}[2]{{#1}\mathrm{e}^{#2}}

\renewcommand{\citeapp}{\cite}

\section{Implementation Details}\label{sec:implementation_details}

\subsection{Pre-training}

\paragraph{Encoders.}  Table~\ref{app:tab:flip:arch} shows the architecture we use. The design follows CLIP~\cite{Radford2021}. Our image encoder involves \mbox{ViT-B}, -L, -H \cite{Dosovitskiy2021}, using the same patch size as in \cite{Dosovitskiy2021} (16 for B and L, 14 for H).
We use global average pooling after the image encoder.
The corresponding text encoder is of a smaller size, following \cite{Radford2021}.
We train ViT-B/-L with 256 TPU-v3 cores, and ViT-H with 512 cores. Table~\ref{app:tab:flip:arch} also shows the model size of the image encoder, text encoder, and the entire model (including output projection layers).

\paragraph{Hyper-parameters.} Our default pre-training configuration is shown in Table~\ref{app:tab:impl_pretrain}. We use the linear learning rate scaling rule \citeapp{Goyal2017}: \textit{lr} = \textit{base\_lr}$\times$batchsize / 256. We observe that using this rule allows us to change the batch size in ablations without extra learning rate search.
The numerical precision we use is float32 by default. We also experimented with bfloat16, but only observed a $\app$1.1$\times$ speedup, which is consistent with the results reported in Google's blog~\footnote{\scriptsize{\url{https://cloud.google.com/blog/products/ai-machine-learning/bfloat16-the-secret-to-high-performance-on-cloud-tpus}}}.

Unmasked tuning, which is a form of pre-training while disabling masking, follows Table~\ref{app:tab:impl_pretrain}, except that we lower the base learning rate to 4e-8 and shorten the warmup schedule to 25.6M samples.

\subsection{ImageNet Classification}

\paragraph{Zero-shot.} 
We follow the prompt engineering in \cite{Radford2021}. Their code provides 80 templates.\footnotemark~We use a subset of 7 templates they recommend; using all 80 templates gives similar results but is slower at inference.

\footnotetext{\scriptsize{\url{https://github.com/openai/CLIP/blob/main/notebooks/Prompt_Engineering_for_ImageNet.ipynb}}}

\paragraph{Linear probing and fine-tuning.} The setting follows~\cite{He2021}. See Table~\ref{app:tab:impl_linear} and Table~\ref{app:tab:impl_finetune}.

\subsection{Zero-shot Retrieval}

We evaluate the performance of zero-shot retrieval on two standard benchmarks: Flickr30K~\cite{Young2014} and COCO~\cite{Lin2014}, respectively with 1K and 5K image-text pairs in their test sets. Following the protocol in CLIP~\cite{Radford2021}, we extract the image and text embeddings from the corresponding encoders and perform retrieval based on the cosine similarities over candidate image-text pairs; no prompt is used.

\subsection{Zero-shot Robustness Evaluation}

In our zero-shot robustness evaluation on the ImageNet-related sets, we use the 7 prompts provided by \cite{Radford2021}, only except in IN-R we use all 80 prompts that are better than the 7 prompts by noticeable margins. The dataset preparation and split follow OpenCLIP \cite{Ilharco2021}.\footnotemark~
In ObjectNet, we follow \cite{Radford2021} to use the class names without prompts. In YTBB, we use the VOC prompts provided by \cite{Radford2021}.

\footnotetext{\scriptsize{\url{https://github.com/LAION-AI/CLIP_benchmark}}}

\begin{table}
	\tablestyle{2pt}{1.1}
	\begin{tabular}{*l | ^c | ^c ^c ^c | ^c ^c ^c | ^c ^c ^c }
		& Embed &    \multicolumn{3}{c|}{Vision Transformer} & \multicolumn{3}{c|}{Text Transformer} & \multicolumn{3}{c}{\# params (M)} \\
		Model & dim  & layers & width & heads  & layers & width & heads & vision & text & total \\ 
		\shline
		B/16  & 512  & 12 & 768 & 12 & 12 & 512 & 8 & 86 & 53 & 141\\
		L/16  & 768  & 24 & 1024 & 16 & 12 & 768 & 12 & 303 & 109 & 414\\
		H/14  & 1024 & 32 & 1280 & 16 & 24 & 1024 & 16 & 631 & 334 & 967\\
	\end{tabular}
	\caption{\textbf{Encoder specifics.}
	}
	\label{app:tab:flip:arch}
\end{table}

\begin{table}[t]
	\tablestyle{6pt}{1.02}
	\scriptsize
	\begin{tabular}{y{96}|y{70}}
		config & value \\
		\shline
		optimizer & AdamW \citeapp{Loshchilov2019} \\
		base learning rate & 4e-6 \\
		weight decay & 0.2 \\
		optimizer momentum & $\beta_1, \beta_2{=}0.9, 0.95$ \cite{Chen2020c} \\
		learning rate schedule & cosine decay \citeapp{Loshchilov2016} \\
		warmup (in samples) & 51.2M (B/L), 256M (H) \\
		numerical precision & float32 \\
	\end{tabular}
	\caption{\textbf{Pre-training setting.}}
	\label{app:tab:impl_pretrain} 
\end{table}

\begin{table}[t]
	\tablestyle{6pt}{1.02}
	\scriptsize
	\begin{tabular}{y{96}|y{68}}
		config & value \\
		\shline
		optimizer & LARS \citeapp{You2017} \\
		base learning rate & 0.01 \\
		weight decay & 0 \\
		optimizer momentum & 0.9 \\
		batch size & 16384 \\
		learning rate schedule & cosine decay \\
		warmup epochs & 10 \\
		training epochs & 90 \\
		augmentation & RandomResizedCrop \\
	\end{tabular}
	\caption{\textbf{Linear probing setting.}
		\label{app:tab:impl_linear}}
\end{table}

\begin{table}[t]
	\tablestyle{6pt}{1.02}
	\scriptsize
	\begin{tabular}{y{96}|y{68}}
		config & value \\
		\shline
		optimizer & AdamW \\
		base learning rate & 5e-5 \\
		weight decay & 0.05 \\
		optimizer momentum & $\beta_1, \beta_2{=}0.9, 0.999$ \\
		layer-wise lr decay \citeapp{Clark2020} & 0.75 \\
		batch size & 1024 \\
		learning rate schedule & cosine decay \\
		warmup epochs & 5 \\
		training epochs & 50 (L/H) \\
		augmentation & RandAug (9, 0.5) \citeapp{Cubuk2020} \\
		label smoothing \citeapp{Szegedy2016a} & 0.1 \\
		mixup \citeapp{Zhang2018a} & 0.8 \\
		cutmix \citeapp{Yun2019} & 1.0 \\
		drop path \citeapp{Huang2016} & 0.2 (L/H) \\
	\end{tabular}
	\caption{\textbf{Fine-tuning setting.}}
	\label{app:tab:impl_finetune}
\end{table}

\subsection{More Zero-shot Datasets}

For the experiments in Table~\ref{tab:more:zeroshot}, we use the prompts provided by \cite{Radford2021}.\footnotemark
~We follow the data preparation scripts provided by \citeapp{Goyal2021a} and \citeapp{Mu2022} and load data using Tensorflow Datasets.
Following \cite{Radford2021}, we report the mean accuracy per class for FGVC Aircraft, Oxford-IIIT Pets, Caltech-101, and Oxford Flowers 102 datasets; we report the mean of top-1 and top-5 accuracy for Kinetics-700, ROC AUC for Hateful Memes, and 11-point mAP for Pascal VOC 2007 Classification; we report top-1 accuracy for the rest of the datasets. We note that the Birdsnap dataset on Internet is shrinking over time and only 1850 test images are available for us (\vs 2149 images tested in \cite{Radford2021}, and 2443 originally).

\footnotetext{\fontsize{6.6pt}{1em}\selectfont{\mbox{\url{https://github.com/openai/CLIP/blob/main/data/prompts.md}}}}

\subsection{Captioning}

We build a sequence-to-sequence encoder-decoder transformer model on top of the ViT image encoder, with 3 encoder layers and 3 decoder layers following \cite{Barraco2022}. Specifically, the ViT image features are first linearly projected to a 384-dimensional sequence and further encoded by a 3-layer transformer encoder (of 384 width and 6 heads). For auto-regressive caption generation, we discard the pre-trained text encoder in FLIP and use a randomly initialized 3-layer transformer decoder (of 384 width and 6 heads) with cross-attention to encoder outputs. The model is trained to predict the next text token using the tokenizer in \cite{Radford2021}.

For simplicity, we supervise the image captioning model only with teacher forcing using a word-level cross-entropy loss \cite{Barraco2022}; we do not use the CIDEr score optimization in~\cite{Barraco2022}. The full model is fine-tuned end-to-end with the AdamW optimizer, a batch size of 256, a learning rate of 1e-4 for newly added parameters, a weight decay of \mbox{1e-2}, a warmup of 15\% iterations, and a cosine decay learning rate schedule. The learning rate for the pre-trained ViT parameters is set to 1e-5 for ViT-L (and 5e-6 for ViT-H). The input image size is 512$\x$512 for ViT-L/16 and 448$\x$448 for ViT-H/14 (to keep the same sequence lengths).

All models are fine-tuned for image captioning on the COCO training split of \cite{Karpathy2015} for 20 epochs. During inference, the image captions are predicted with auto-regressive decoding, and we report their performance on the COCO test split of \cite{Karpathy2015} under different metrics.

To evaluate how the COCO-trained models generalize to novel objects, we evaluate these models directly on the nocaps \cite{Agrawal2019} validation set, with no further fine-tuning.

\subsection{Visual Question Answering} 

In our VQA experiments, we follow the architecture described in \cite{Dou2022}. Specifically, the VQA task is casted as a classification problem over all answer classes.
The input images are encoded by the ViT encoders.
The input questions are encoded by a pre-trained RoBERTa text encoder \citeapp{Liu2019a}, following the practice in \cite{Dou2022}. A multimodal fusion Transformer (4 layers, 768-d, 12 heads, with merged attention \cite{Dou2022}) is applied to combine the image and text representations. A two-layer MLP is applied on the class token of the fusion module to obtain the VQA output \cite{Dou2022}.

We fine-tune the VQA model end-to-end. The loss function is a binary sigmoid loss using soft scores \citeapp{Teney2018}.
We use a batch size of 256, a learning rate of 1e-4 for randomly initialized parameters, and a learning rate of 1e-5 (ViT-L) or 5e-6 (ViT-H) for the pre-trained ViT parameters. We use a weight decay of \mbox{1e-2}, a warmup of 15\% of iterations, and a cosine decay learning rate schedule. The input image size is 512$\x$512 for ViT-L/16 and 448$\x$448 for ViT-H/14.

All models are fine-tuned for 20 epochs on the VQAv2 train+val set, with additional question-answer pairs from Visual Genome~\citeapp{Krishna2017}, following \citeapp{Teney2018}. We report results on the test-dev split from the evaluation server.

{\fontsize{8.6pt}{10.32pt}\selectfont
\bibliographystyle{ieee_fullname}
\bibliography{flip}
}

\end{document}